\pdfoutput=1

\documentclass[11pt]{article}
\usepackage[table]{xcolor}

\usepackage[final]{acl}

\usepackage{times}
\usepackage{latexsym}

\usepackage[T1]{fontenc}

\usepackage[utf8]{inputenc}

\usepackage{microtype}

\usepackage{inconsolata}

\usepackage{graphicx}

\usepackage{hyperref}
\usepackage{url}
\usepackage{booktabs}
\usepackage{orcidlink}
\usepackage{multirow}
\usepackage{xspace}
\usepackage{capt-of}
\usepackage{color}
\usepackage{enumitem}
\usepackage{wrapfig}
\usepackage{comment}
\usepackage{amsmath}
\usepackage{amsfonts}

\def\checkmark{\tikz\fill[scale=0.25](0,.35) -- (.25,0) -- (1,.7) -- (.25,.15) -- cycle;}

\definecolor{lightorange}{RGB}{255, 204, 153}

\title{Test-Time Consistency in Vision Language Models}

\author{Shih-Han Chou\thanks{Equal Contribution, order determined by coin flip.}$^{1,2}$, Shivam Chandhok$^{\ast1,2}$, James J. Little$^{1}$, Leonid Sigal$^{1,2,3}$\\
$^{1}$Department of Computer Science, University of British Columbia \\ 
$^{2}$Vector Institute for AI \hspace{2mm}
$^{3}$Canada CIFAR AI Chair\\
\texttt{\{shchou75, chshivam, little, lsigal\}@cs.ubc.ca} \\
}

\begin{document}
\maketitle
\begin{abstract}
Vision-Language Models (VLMs) have achieved impressive performance across a wide range of multimodal tasks, yet they often exhibit inconsistent behavior when faced with semantically equivalent inputs—undermining their reliability and robustness. Recent benchmarks, such as MM-R$^3$, highlight that even state-of-the-art VLMs can produce divergent predictions across semantically equivalent inputs, despite maintaining high average accuracy. Prior work addresses this issue by modifying model architectures or conducting large-scale fine-tuning on curated datasets. In contrast, we propose a simple and effective \emph{test-time consistency framework} that enhances semantic consistency \emph{without supervised re-training}.
Our method is entirely \emph{post-hoc}, model-agnostic, and applicable to any VLM with access to its weights. Given a single test point, 
we enforce consistent predictions via two complementary objectives: (i) a \textbf{Cross-Entropy Agreement Loss} that aligns predictive distributions across semantically equivalent inputs, and (ii) a \textbf{Pseudo-Label Consistency Loss} that draws outputs toward a self-averaged consensus. Our method is \textit{plug-and-play}, and leverages information from a single test-input itself to improve consistency. Experiments on the MM-R$^3$ benchmark show that our framework yields substantial gains in consistency across state-of-the-art models, establishing a new direction for inference-time adaptation in multimodal learning.
\end{abstract}

\section{Introduction}

Vision-Language Models (VLMs)~\cite{liu2024improved,liu2024llavanext,wang2024qwen2,hurst2024gpt} have achieved impressive performance across a wide range of multimodal tasks, including visual question answering~\cite{VQA}, captioning~\cite{lin2014microsoft, sharma2018conceptual, chen2015microsoft}, and reasoning~\cite{johnson2017clevr,zellers2019vcr}. While existing evaluations predominantly focus on accuracy, a growing body of work highlights a critical shortcoming: \emph{semantic inconsistency}. That is, VLMs producing divergent outputs when prompted with semantically equivalent inputs—undermining their reliability, interpretability, and deployment in high-stakes settings.
\begin{figure}[t]
  \centering
  \includegraphics[width=.85\columnwidth]{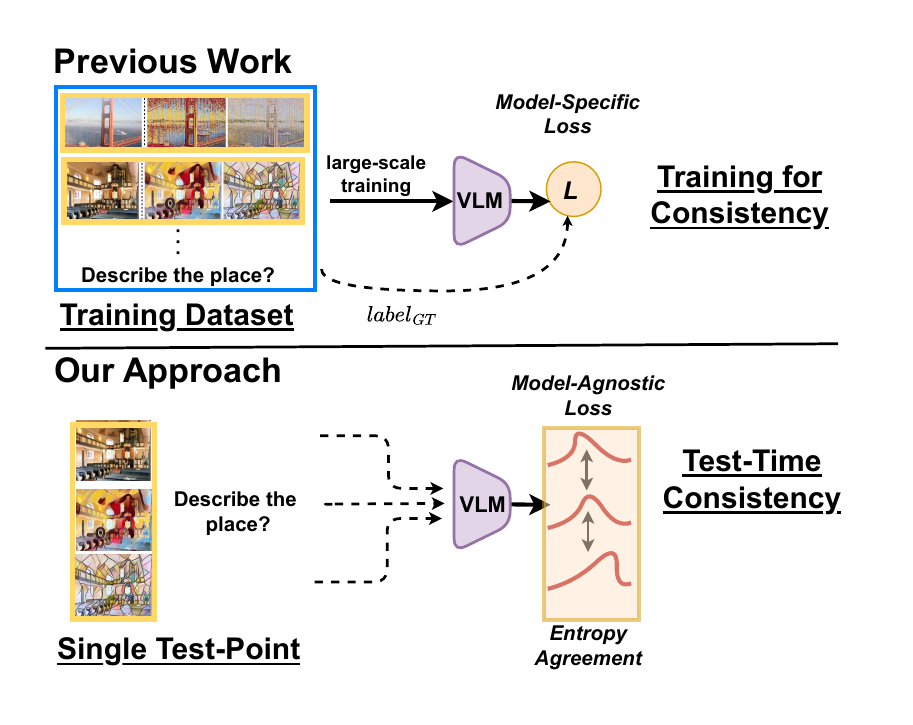}
  \vspace{-.5em}
  \caption{
  Comparison between \textbf{training-time} (top) and our \textbf{test-time} (bottom) consistency frameworks. While prior work ({\em e.g.}, ~\cite{chou2024mm}) needs large-scale supervised fine-tuning with curated dataset to enforce consistency, our method operates entirely post-hoc by adapting to a single test point with few gradient steps at test-time\vspace{-0.25in}
  }
  \label{fig:intro}
\end{figure}

Despite achieving strong accuracy on vision-language tasks, modern VLMs often yield inconsistent responses when presented with semantically equivalent variants of a test query. The MM-R$^3$ benchmark, recently proposed in~\citet{chou2024mm},  highlights this issue by evaluating models under three types of controlled perturbations: \textit{question rephrasing}, \textit{image restyling}, and \textit{context masking}. Results reveal that even state-of-the-art models show significant variance across these conditions, illustrating that high accuracy does not imply semantic consistency—a fundamental prerequisite for robust multimodal reasoning.
While~\citet{chou2024mm} address consistency via adapter-based fine-tuning, their approach requires intrusive architectural modifications and access to a sizable training dataset. 

In contrast, we study the problem of improving \textit{\textbf{consistency at test time}}—using only the test input itself, without any access to model internals, training data, or loss functions. Unlike prior methods that rely on supervised re-training, additional curated datasets, or adapter insertion \cite{chou2024mm}, our framework makes \emph{no assumptions about training} and operates entirely at inference. This is especially important in real-world where retraining is infeasible due to proprietary models, limited compute, or lack of access to original training data.

We propose a simple, general-purpose test-time consistency framework that can be applied to any probabilistic vision-language model (VLM) in a \textit{plug-and-play} fashion. Our method leverages semantically equivalent variants of a given input and encourages agreement among the resulting predictions using two lightweight objectives: (1) a Cross-Entropy Agreement Loss, which penalizes divergence in output distributions, and (2) a Pseudo-Label Consistency Loss, which aligns predictions toward a consensus output. Crucially, our method requires no modifications to the model architecture and operates entirely at inference time. Intuitively, it encourages the model to generate invariant predictions across different linguistic and visual realizations of the same query, thereby promoting robustness and semantic stability.

Importantly, our method adapts model behavior at test time—even for a \emph{single} input—by utilizing the information embedded in that input’s semantic variations. This departs from training-centric paradigms and instead exploits the rich signal present in the test data itself, which prior work often overlooks. Because our framework requires no access to original training data, loss functions, or model internals, and avoids retraining or auxiliary supervision, it can be seamlessly integrated into any VLM pipeline regardless of architecture.

We evaluate our approach on the standard MM-R$^3$ benchmark and demonstrate substantial improvements in consistency across multiple open-source and proprietary VLMs. Our results show that even strong models benefit from targeted inference-time regularization, and we advocate for consistency—as well as accuracy—to be a central design goal in future multimodal learning systems.

Our work makes the following \textbf{contributions}:
\begin{itemize}[leftmargin=*,noitemsep]
\vspace{-.8em}
    \item We address the underexplored problem of \textit{test-time consistency} in VLMs by proposing a simple, model-agnostic framework that improves consistency in VLM response across semantically equivalent inputs. Our method operates entirely \emph{post-hoc}—requiring only access to model weights and using information derived solely from a \emph{single test input}. It requires no training dataset, no access to original loss functions or training procedures, and no supervised retraining—making it broadly applicable across models and practical deployment settings.
    
    \item Our framework leverages two complementary objectives: (1) a Cross-Entropy Agreement Loss to reduce divergence among predictions on perturbed inputs, and (2) a Pseudo-Label Consistency Loss to align predictions towards a consensus output. Unlike prior work focused on training-time consistency or fine-tuning, our method makes no assumptions about model training, and is \textit{plug-and-play}, requiring only access to model outputs. 

    \item We show that our framework significantly improves consistency across linguistic and visual perturbations in the MM-R$^3$ benchmark without retraining or architectural changes. Our results highlight that even a single test point contains valuable signal that can be used to adapt model behavior at inference, offering a new paradigm for robust multimodal reasoning.
\end{itemize}

\section{Related Works}

\paragraph{Consistency in Vision-Language Models.}  
While existing evaluations of Vision-Language Models (VLMs) predominantly focus on accuracy, a growing body of work highlights a critical shortcoming: \emph{semantic inconsistency} ~\cite{chou2024mm}. That is, VLMs often produce divergent outputs when prompted with semantically equivalent inputs—undermining their reliability, interpretability, and applicability in high-stakes settings.
The MM-R$^3$ benchmark~\cite{chou2024mm} systematically investigates this issue, introducing a suite of perturbation-based evaluations across rephrased questions, stylized images, and masked contexts. Their results show that even state-of-the-art VLMs exhibit significant inconsistency across these settings, despite high accuracy—revealing a fundamental gap between correctness and stable reasoning.

While prior efforts to improve consistency ~\cite{chou2024mm} typically focus on modifying training objectives, leveraging larger models, or fine-tuning on curated data, these approaches are computationally intensive and often impractical. In contrast, we address this challenge from a test-time perspective, proposing a lightweight, post-hoc framework that improves consistency without retraining or access to labels.

\paragraph{Test-Time Adaptation.}
Test-time adaptation methods have progressed from entropy-based confidence maximization to efficient modular tuning. MEMO~\cite{zhang2022memo} improves robustness by enforcing confident and consistent predictions across augmented test-time views. Test-Time Prompt Tuning~\cite{shu2022test} adapts CLIP by optimizing prompts at inference to better match shifted distributions. MedAdapter~\cite{shi-etal-2024-medadapter} steers pretrained LLMs toward domain-specific tasks by updating small adapter modules without full retraining. LoRA-TTT~\cite{kojima2025lora} further reduces adaptation cost by fine-tuning low-rank adapters at test time. \citet{karmanov2024efficient} proposed a lightweight VLM adaptation strategy that freezes the core model and updates only a small projection head via entropy minimization.
While most of these works focus on improving accuracy via test-time optimization, our work targets a complementary and underexplored axis: \textit{semantic consistency}. Unlike methods that require retraining, or architectural modifications, our approach is entirely post-hoc, model-agnostic, and leverages information from a single test input—making it lightweight, scalable, and broadly applicable.
\paragraph{Pseudo-Labeling and Self-Training.}
Pseudo-labeling has been widely used in semi-supervised learning~\cite{yarowsky-1995-unsupervised, lee2013pseudo, berthelot2019mixmatch, sohn2020fixmatch, zhang2021flexmatch}, often paired with augmentations or confidence thresholds. In vision-language models, it has been employed to generate pseudo-captions~\cite{yang2022self}, region–phrase alignments~\cite{chou2022semi}, and visual-language prototypes~\cite{ali2025dpa}. We adopt a test-time variant of pseudo-labeling, aggregating model outputs across perturbed inputs into a self-consistent consensus—encouraging stability without requiring external supervision or retraining.

\paragraph{Entropy-Based Adaptation.}
Entropy minimization has been a foundational strategy for improving robustness under distribution shift. Grandvalet and Bengio~\cite{grandvalet2004semi} introduced it as a regularization objective for unlabeled data, and TENT~\cite{wang2021tent} applied it for test-time adaptation by optimizing batch norm parameters. MEMO~\cite{zhang2022memo} improved on this by combining entropy minimization with multi-view consistency during inference. Extensions to large models include entropy-guided generation in LLMs~\cite{kuhn2023semantic, farquhar2024detecting} and efficient test-time tuning for vision-language models by updating only lightweight projection heads~\cite{karmanov2024efficient}. Our method builds on this line of work by extending entropy-based objectives to open-ended multimodal settings—not to improve accuracy, but to enhance semantic consistency under perturbations, an underexplored yet practically important aspect of reliability and interpretability in multimodal reasoning.

\section{Approach}
\subsection{Problem Setting}

We follow the procedure and settings defined in the MM-R$^3$ benchmark to evaluate consistency under diverse semantic variations. Given a test input $\mathbf{x} = (I, Q)$, the benchmark provides $K$ semantically equivalent variants $(I_k, Q_k)$ constructed via:
\begin{itemize}[leftmargin=*,noitemsep]
\vspace{-.8em}
    \item \underline{\em Question Rephrasing}: Paraphrased variants of $Q$ generated using a language model keeps $I$ fixed.
    \item \underline{\em Image Restyling}: Stylized versions of $I$ using neural style transfer ({\em e.g.}, Mosaic, Candy, Undie, and Grayscale) with $Q$ not altered.
    \item \underline{\em Context Reasoning}: Variants of $I$ with different occlusions applied to a specific object region, while keeping $Q$ once again fixed.
\vspace{-.8em}
\end{itemize}

Each perturbed pair $(I_k, Q_k)$ is passed through the VLM to obtain response distributions:
\vspace{-.8em}
\begin{equation}
\vspace{-.5em}
    \mathbf{p}_k = \mathrm{\tt VLM}(I_k, Q_k), \quad \text{for } k = 1, \ldots, K
\end{equation}

\subsection{Method}

\paragraph{Overview.} We propose a lightweight, test-time strategy to improve the semantic consistency of Vision-Language Models (VLMs) by encouraging agreement across semantically equivalent variants of a single test input. Our method operates entirely post-hoc and leverages only the information present in the given test example. It performs a small number of inference-time updates (typically 1–4 steps), requiring no access to training data, ground-truth labels, or model internals.

Our approach combines two complementary objectives: (1) a \textbf{Cross-Entropy Agreement Loss} that aligns token-level output distributions across perturbed inputs, and (2) a \textbf{Pseudo-Label Consistency Loss} that enforces convergence toward a stable, consensus output prediction. These objectives guide the model to become more consistent at inference, without altering its original architecture or parameters via supervised training.

\begin{figure}[t]
  \centering
  \includegraphics[width=\columnwidth]{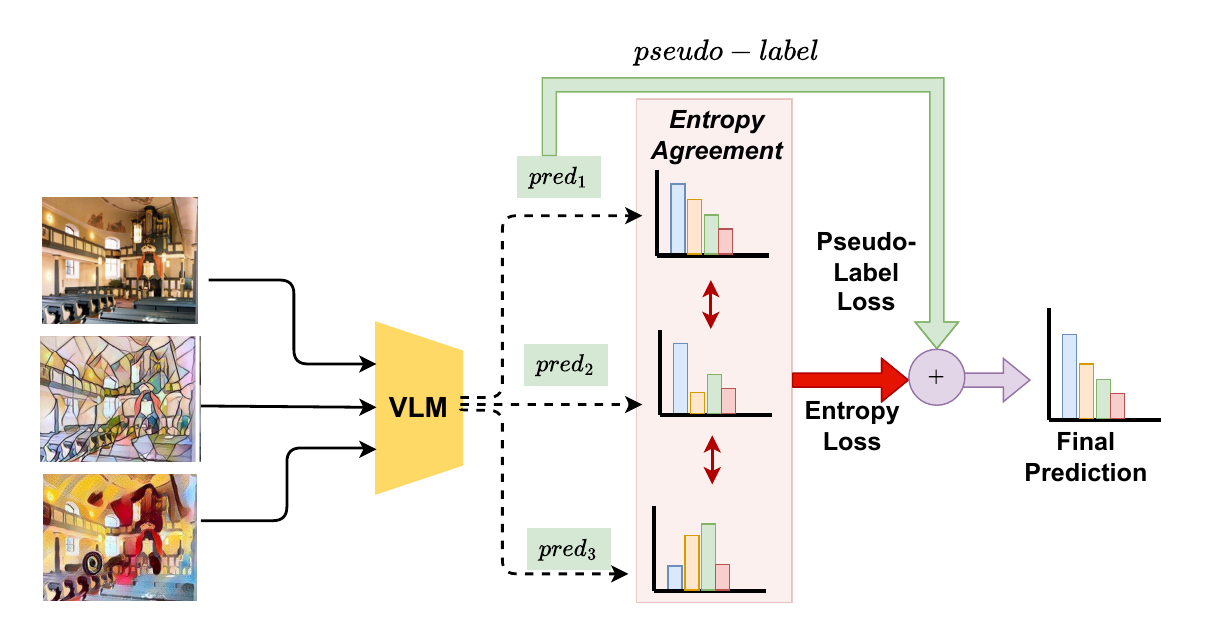}
  \caption{\textbf{Overview of our test-time consistency framework}. Given a test input with semantically equivalent input variants ({\em e.g.}, restyled images), we forward them through a pretrained VLM to obtain predictions. Two complementary objectives are used to improve consistency: (1) Cross-Entropy Agreement Loss, which aligns token-level output distributions across variants, and (2) Pseudo-Label Consistency Loss, which encourages agreement with a consensus pseudo-label. The model is updated with few (1-4) steps using gradients from these objectives, enabling consistent final predictions without access to training data or model internals.
  }
  \vspace{-1em}
  \label{fig:method}
\end{figure}
\subsection{Cross-Entropy Agreement Loss}

To promote consistency across semantically equivalent input variants, we introduce a Cross-Entropy Agreement Loss that aligns their token-level output distributions. Given a test input, we generate VLM output for $K$ perturbed variants and obtain token-level logits for each through a forward pass.

Let $\mathbf{z}_k^j \in \mathbb{R}^V$ denote the logits over the vocabulary $V$ at output token position $j$ of the VLM response for the $k$-th input variant. Let $L_k$ be the total number of valid output tokens in response for that variant. We compute the average logits across the decoded sequence for each variant $k$:
\vspace{-.8em}
\begin{equation}
\vspace{-.8em}
\bar{\mathbf{z}}_k = \frac{1}{L_k} \sum_{j=1}^{L_k} \mathbf{z}_k^j
\end{equation}

We then apply softmax to obtain the normalized token distribution:
\vspace{-.8em}
\begin{equation}
\vspace{-.5em}
    \mathbf{p}_k = \mathrm{softmax}(\bar{\mathbf{z}}_k)
\end{equation}

The agreement loss is defined as the average of all pairwise symmetric cross-entropies across the $K$ output distributions:
\vspace{-.5em}
\begin{equation}
\vspace{-.5em}
\small
    \mathcal{L}_{\mathrm{CE}} = \frac{2}{K(K-1)} \sum_{i < j} \mathrm{CE}(\mathbf{p}_i, \mathbf{p}_j) + \mathrm{CE}(\mathbf{p}_j, \mathbf{p}_i)
\end{equation}

This loss encourages alignment of the global output tokens across $K$ input variants while ensuring the model's generation is \emph{distributionally consistent}, even if wording or phrasing changes.

\subsection{Pseudo-Label Consistency Loss}

To complement distributional alignment, we introduce a Pseudo-Label Consistency Loss that enforces consistency at the output level by aligning each variant’s predicted sequence to a common consensus output prediction.

Let $\{ \mathbf{y}_1, \dots, \mathbf{y}_K \}$ be the decoded textual outputs from the $K$ semantically equivalent input variants, generated using greedy decoding. To compute a consensus label, we define a string similarity function $\text{sim}(\cdot, \cdot)$ based on normalized Levenshtein distance ({\em e.g.}, token set ratio). We cluster the $K$ output responses by assigning two responses $\mathbf{y}_i$ and $\mathbf{y}_j$ to the same cluster if
\vspace{-.5em}
\begin{equation}
\vspace{-.5em}
    \text{sim}(\mathbf{y}_i, \mathbf{y}_j) \geq \tau,
\end{equation}
where $\tau \in [0,1]$ is a fixed similarity threshold ({\em i.e.}, $\tau = 0.85$). Among all clusters, we identify the largest one, and from within it, select the most frequent response as the pseudo-label:
\vspace{-.5em}
\begin{equation}
\vspace{-.5em}
\hat{\mathbf{y}}_{\text{pseudo}} = \text{mode} \left( \mathcal{C}_{\max} \right),
\end{equation}
where $\mathcal{C}_{\max}$ is the largest similarity-based cluster.

We then tokenize $\hat{\mathbf{y}}_{\text{pseudo}}$ and use it as the supervision target for all $K$ variants. Let $\mathbf{p}_k$ denote the token-level predicted distribution from variant $k$ ({\em i.e.}, the model’s output logits after softmax). The \textit{Pseudo-Label Consistency Loss} is defined as:
\vspace{-.5em}
\begin{equation}
\vspace{-.5em}
    \mathcal{L}_{\mathrm{PL}} = \frac{1}{K} \sum_{k=1}^{K} \mathrm{CE}(\hat{\mathbf{y}}_{\text{pseudo}}, \mathbf{p}_k),
\end{equation}
where $\mathrm{CE}(\cdot, \cdot)$ denotes the cross-entropy loss between pseudo-label tokens and predicted distribution. This loss encourages all variants to converge to the dominant semantic response, enhancing answer-level consistency of perturbed inputs.

\vspace{0.5em}
\noindent\textbf{Complementarity of Losses.} The Cross-Entropy Agreement Loss encourages \textit{token-level alignment} by smoothing output distributions across input variants, while the Pseudo-Label Consistency Loss enforces \textit{prediction-level convergence} by aligning decoded outputs with a dominant consensus response. Together, these losses regularize both the internal generation process and final output, yielding improved semantic consistency at test time using only the information in single-test point without modifying the underlying model.

\subsection{Final Objective and Inference}
\label{Inference}

Given a test input with $K$ semantically equivalent variants $(I_k, Q_k)$ ({\em e.g.}, via question rephrasing, image restyling, or context masking), we adapt the model using gradients from two complementary objectives: the Cross-Entropy Agreement Loss $\mathcal{L}_{\mathrm{CE}}$ and the Pseudo-Label Consistency Loss $\mathcal{L}_{\mathrm{PL}}$.

The total loss at each update step is computed as a weighted sum:
\vspace{-.5em}
\begin{equation}\label{overall_loss}
\vspace{-.5em}
    \mathcal{L}_{\mathrm{total}} = \alpha \cdot \mathcal{L}_{\mathrm{CE}} + \beta \cdot \mathcal{L}_{\mathrm{PL}},
\end{equation}
where $\alpha$ and $\beta$ are hyperparameters balancing distributional agreement and semantic convergence. We optimize this objective for a small number of gradient-based updates—typically between $1$ and $4$—using only the current test example, without access to any labeled data or training corpus.

\paragraph{Adaptive Step Selection}

Different test inputs may benefit from different numbers of adaptation steps—while some improve with a few updates, others may degrade due to over-adaptation. To address this, we introduce an adaptive mechanism that dynamically selects the optimal number of steps for each test point.

After each update step \( t \in \{0, 1, \dots, T\} \), we decode the model’s output responses for the \( K \) input variants. To assess internal consistency, we compute the average pairwise token-set similarity—based on normalized Levenshtein distance—among the \( K \) decoded answers and the previously generated pseudo-label (used in the Pseudo-Label Consistency Loss). The similarity score for step \( t \) is given by:
\vspace{-.5em}
\begin{equation}
\vspace{-.5em}
    \text{score}_t = \frac{1}{K(K-1)} 
    \sum_{i < j} \text{sim}(a_i^t, a_j^t),
\end{equation}
where \( a_i^t \) and \( a_j^t \) denote either one of the \( K \) decoded answers or the shared pseudo-label. The step \( t^* \) with the highest score is selected as the final output, reflecting the most consistent model behavior during adaptation.

\vspace{-.5em}
\begin{equation}
\vspace{-.5em}
    t^* = \arg\max_t \text{score}_t.
\end{equation}

This selection mechanism is fully unsupervised and relies solely on model outputs \textit{without using ground-truth annotations}. It enables per-instance, test-time adaptation that is both robust and efficient, ensuring that predictions remain semantically consistent while avoiding over-updating.
\paragraph{Method Variants.}

We report results for two variants of our method. In the first, we use a fixed number of adaptation steps ($T=2$) for all samples, which we refer to as \textbf{Test-time (constant $T$)}. In the second, we use the adaptive step selection mechanism described above to dynamically choose the optimal number of updates per input. We refer to this variant as \textbf{Test-time (adapt. $T$)}. This comparison allows us to assess the trade-offs between simplicity and input-specific adaptivity.

\section{Experiments}
\paragraph{Dataset.}

We evaluate our method on the standard MM-R$^3$ dataset~\cite{chou2024mm}, which consists of three test-time consistency tasks: \textbf{question rephrasing}, \textbf{image restyling}, and \textbf{context reasoning}. The \textit{question rephrasing} task assesses whether VLMs produce consistent answers to semantically equivalent questions phrased differently. The \textit{image restyling} task evaluates consistency under visual domain shifts by presenting stylized versions of the image. The \textit{context reasoning} task tests the model’s ability to reason under partial occlusion. Our evaluations are conducted on MM-R$^3$ test set.

\paragraph{Models.}
We evaluate our method on widely used state-of-the-art open-source Vision-Language Models (VLMs). Specifically, 
LLaVA 1.5M~\cite{liu2024improved} (\texttt{llava-v1.5-7b} version), LLaVA-Next~\cite{liu2024llavanext} (\texttt{llava-v1.6-mistral-7b} checkpoint), and Qwen2-VL~\cite{wang2024qwen2} (\texttt{Qwen2-VL-7B-Instruct} variant). These are all strong VLM models, with Qwen2-VL broadly considered the strongest among the three, 

Our choice of these models is motivated by the fact that these models are widely used as foundations for downstream applications and frequently serve as initialization points for developing more advanced VLMs. Since our method involves modifying model parameters at test time, we restrict our evaluation to open-source models and exclude proprietary systems such as GPT-4V or Gemini. Evaluating on these representative models enables us to assess the generality, practical utility, and broader impact of test-time consistency improvements across VLMs.

\noindent\textbf{Implementation Details.} Please refer to the Appendix~\ref{appdex:implementation}.

\paragraph{Evaluation Metrics.}  
Since VLM responses are open-ended and linguistically diverse, we adopt evaluation metrics similar to those introduced in MM-R$^3$~\cite{chou2024mm}, in order to capture both correctness and consistency. We briefly introduce the core evaluation metrics used to assess correctness and consistency; full metric definitions and implementation details are provided in Appendix.

\begin{itemize}[leftmargin=*,noitemsep]
\vspace{-.5em}
\item \textbf{Accuracy ($\mathbf{Acc}$):} Measures correctness using fuzzy string matching, accounting for minor lexical variations. A similarity threshold of 85 is used to determine a match.

\item \textbf{Similarity with Ground Truth ($S_{\text{GT}}$):} Computes semantic similarity between the model’s response and the reference answer using BERT sentence embeddings, offering a more flexible alternative to exact match.

\item \textbf{Consistency Accuracy ($\text{Con}$):} Evaluates semantic agreement across responses to semantically equivalent inputs. Responses are considered consistent if their pairwise similarity exceeds a threshold of 0.7.

\item \textbf{Consistency Similarity ($S_{\text{C}}$):} Computes the average pairwise similarity across all response variations, providing a smoother measure of output invariance.

\item \textbf{Overall Score ($O_{all}$):} The harmonic mean of correctness and consistency metrics.
\vspace{-.2em}
\begin{equation}
\small
H_{mean}(mean(\mathbf{Acc},\mathbf{S_{GT}}),mean(\mathbf{Con},\mathbf{S_C})).
\vspace{-1em}
\end{equation}

We use the harmonic mean to emphasize models that are balanced in both accuracy and consistency, as it penalizes performance when either component is low. This provides a unified measure of overall model quality.
\end{itemize}

\begin{table}[!t]
\scriptsize
\begin{center}
\caption{\textbf{Overall results.} We highlight our approach in orange color and the overall results in gray color. The best-performing method is in bold for each models.}
\vspace{-1em}
\begin{tabular}{c|l|c|c|c|c|c}
    \toprule
    & \textbf{Models} & $\mathbf{Acc}$ & $\mathbf{S_{GT}}$ & $\mathbf{Con}$ & $\mathbf{S_{C}}$ & $O_{all}$ \\
    \midrule
    \parbox[t]{1mm}{\multirow{9}{*}{\rotatebox[origin=c]{90}{\parbox{2.6cm}{\centering\textbf{Question Rephrasing}}}}} & LLaVa 1.5M & 36.18 & 62.96 & 48.55 & 64.10 & \cellcolor{gray!25}{52.73} \\ 
    & \cellcolor{lightorange!25}{~ + Constant $T$} & 38.00 & 65.05 & 77.67 & 84.65 & \cellcolor{gray!25}{63.03} \\
    & \cellcolor{lightorange!25}{~ + Adapt. $T$} & 39.58 & 65.10 & 79.11 & 86.10 & \cellcolor{gray!25}{\textbf{64.08}}\\
    & LLaVa-Next & 42.89 & 64.89 & 49.18 & 65.69 & \cellcolor{gray!25}{55.61} \\
    & \cellcolor{lightorange!25}{~ + Constant $T$} & 44.48 & 68.74 & 83.39 & 88.47 & \cellcolor{gray!25}{68.25} \\
    & \cellcolor{lightorange!25}{~ + Adapt. $T$} & 44.74 & 68.67 & 85.18 & 89.92 & \cellcolor{gray!25}{\textbf{68.83}} \\
    & Qwen2-VL & 66.72 & 79.69 & 65.78 & 76.16 & \cellcolor{gray!25}{72.07} \\
    & \cellcolor{lightorange!25}{~ + Constant $T$} & 70.79 & 82.66 & 90.44 & 93.52 & \cellcolor{gray!25}{83.66} \\
    & \cellcolor{lightorange!25}{~ + Adapt. $T$} & 72.14 & 83.27 & 93.6 & 95.64 & \cellcolor{gray!25}{\textbf{85.33}}\\
    \midrule
    \midrule
    \parbox[t]{1mm}{\multirow{9}{*}{\rotatebox[origin=c]{90}{\parbox{2.6cm}{\centering\textbf{Image Restyling}}}}} & 
    LLaVa 1.5M & 9.61 & 34.85 & 18.96 & 56.91 & \cellcolor{gray!25}{28.03}\\
    & \cellcolor{lightorange!25}{~ + Constant $T$} & 12.09 & 35.62 & 20.14 & 59.01 & \cellcolor{gray!25}{29.77}\\
    & \cellcolor{lightorange!25}{~ + Adapt. $T$} & 17.94 & 40.15 & 33.90 & 64.46 & \cellcolor{gray!25}{\textbf{36.52}}\\
    & LLaVa-Next & 17.57 & 41.47 & 55.34 & 71.36 & \cellcolor{gray!25}{40.27} \\
    & \cellcolor{lightorange!25}{~ + Constant $T$} & 18.99 & 42.49 & 88.25 & 91.25 & \cellcolor{gray!25}{45.80} \\
    & \cellcolor{lightorange!25}{~ + Adapt. $T$} & 18.71 & 42.52 & 91.85 & 93.16 & \cellcolor{gray!25}{\textbf{46.00}} \\
    & Qwen2-VL & 21.13 & 39.25 & 61.67 & 75.85 & \cellcolor{gray!25}{41.96} \\
    & \cellcolor{lightorange!25}{~ + Constant $T$} & 22.60 & 42.32 & 98.30 & 98.97 & \cellcolor{gray!25}{48.85} \\
    & \cellcolor{lightorange!25}{~ + Adapt. $T$} & 
    22.58 & 46.40 & 99.14 & 99.45 & \cellcolor{gray!25}{\textbf{51.20}}\\
    \midrule
    \midrule
    \parbox[t]{1mm}{\multirow{9}{*}{\rotatebox[origin=c]{90}{\parbox{2.6cm}{\centering\textbf{Context Reasoning}}}}} & 
    LLaVa 1.5M & 16.11 & 42.69 & 65.64 & 75.08 & \cellcolor{gray!25}{41.47} \\
    & \cellcolor{lightorange!25}{~ + Constant $T$} & 22.88 & 49.49 & 88.89 & 93.45 & \cellcolor{gray!25}{51.81} \\
    & \cellcolor{lightorange!25}{~ + Adapt. $T$} & 31.04 & 55.14 & 72.11 & 81.90 & \cellcolor{gray!25}{\textbf{55.26}} \\
    & LLaVa-Next & 30.24 & 27.43 & 32.11 & 58.44 & \cellcolor{gray!25}{35.23} \\
    & \cellcolor{lightorange!25}{~ + Constant $T$} & 32.50 & 50.84 & 89.91 & 90.16 & \cellcolor{gray!25}{56.97} \\
    & \cellcolor{lightorange!25}{~ + Adapt. $T$} & 32.29 & 53.85 & 95.24 & 96.66 & \cellcolor{gray!25}{\textbf{59.45}}\\
    & Qwen2-VL & 29.09 & 40.03 & 34.58 & 53.70 & \cellcolor{gray!25}{38.77} \\
    & \cellcolor{lightorange!25}{~ + Constant $T$} & 29.60 & 50.11 & 91.17 & 91.75 & \cellcolor{gray!25}{55.52} \\
    & \cellcolor{lightorange!25}{~ + Adapt. $T$} & 30.42 & 53.00 & 99.53 & 99.66 & \cellcolor{gray!25}{\textbf{58.80}}\\
    \bottomrule
\end{tabular}
\vspace{-2em}
\label{tab:main}
\end{center}
\end{table}

\vspace{-1em}
\subsection{Main Results}
\paragraph{Overview.}  
Table~\ref{tab:main} presents the performance of our test-time consistency framework across three tasks in the MM-R$^3$ benchmark. 
We report results for each base model with two variants: \textbf{Test-time (constant $T$)} and \textbf{Test-time (adaptive $T$)}. 
Across all tasks, our method consistently improves semantic consistency and overall performance, with the adaptive variant yielding the best results.

\paragraph{Question Rephrasing.}  
In the rephrasing task, our adaptive test-time method yields substantial gains in consistency and overall score across all three models while preserving accuracy. For instance, on LLaVA-1.5M, $O_{all}$ improves from 52.73 (base) to 64.08, with consistency rising from 48.55 to 79.11 and $\textbf{Acc}$ increasing from 36.18 to 39.58. LLaVA-Next and Qwen2-VL also show notable gains, with the adaptive variant achieving the best $O_{all}$ for each model: 68.83 and 85.33, respectively. These results validate the ability of our method to enforce semantic invariance across linguistic perturbations without reducing accuracy. 

\paragraph{Image Restyling.}  
This task poses a significant domain shift challenge due to stylized visual inputs. Our method leads to especially large improvements in consistency for all models. On LLaVA-Next, consistency improves from 55.34 (base) to 91.85 (Test-time) and further to 91.85 (Adaptive), with $O_{\text{all}}$ reaching 46.00. Qwen2-VL sees the highest performance overall, with the adaptive variant achieving $O_{all}$ = 51.20 and nearly perfect consistency (99.14). These results demonstrate the robustness of our framework under visual perturbations.

\paragraph{Context Reasoning.}  
Our approach also improves model behavior in the context reasoning task, which requires stable answers under partial information. Our method delivers both higher consistency and improved accuracy. Specifically, LLaVA-1.5M shows a dramatic gain in $O_{all}$ from 41.47 to 55.26 (adaptive), while LLaVA-Next reaches the highest score of 59.45. Interestingly, even though Qwen2-VL starts from a stronger baseline, our method boosts its $O_{all}$ to 58.80 and consistency to 99.53. These results suggest that test-time consistency not only stabilizes outputs but also improves factual grounding under ambiguity.

\begin{table}[!t]
\scriptsize
\begin{center}
\caption{\textbf{Comparison of our approach with supervised fine-tuned model on LLaVa 1.5M model.}}
\vspace{-1em}
\begin{tabular}{l|ccccc}
    \toprule
    \textbf{Models} & $\mathbf{Acc}$ & $\mathbf{S_{GT}}$ & $\mathbf{Con}$ & $\mathbf{S_{C}}$ & $O_{all}$  \\
    \midrule
    \multicolumn{5}{l}{\textbf{Question Rephrasing}}\\
    \midrule
    LLaVa 1.5M & 36.18 & 62.96 & 48.55 & 64.1 & \cellcolor{gray!25}{52.73}\\
    ~ + Finetuning & & & & & \cellcolor{gray!25}\\
    ~~~~ {\tiny \cite{chou2024mm}} & \multirow{-2}{*}{{42.55}} & \multirow{-2}{*}{{69.03}} & \multirow{-2}{*}{{63.79}} & \multirow{-2}{*}{{75.83}} & \multirow{-2}{*}{\cellcolor{gray!25}{{62.02}}}\\
     ~ + Adapt. $T$ & 39.58 & 65.10 & 79.11 & 86.10 & \cellcolor{gray!25}{\textbf{64.08}}\\
    \midrule
    \midrule
    \multicolumn{5}{l}{\textbf{Image Restyling}}\\
    \midrule
    LLaVa 1.5M & 9.61 & 34.85 & 18.96 & 56.91 & \cellcolor{gray!25}{28.03}\\
    ~ + Finetuning & & & & & \cellcolor{gray!25}\\
    ~~~~ {\tiny \cite{chou2024mm}} &
    \multirow{-2}{*}{25.45} &
    \multirow{-2}{*}{50.67} &
    \multirow{-2}{*}{50.94}  & 
    \multirow{-2}{*}{66.06} &
    \multirow{-2}{*}{\cellcolor{gray!25}{\textbf{46.11}}} \\
    ~ + Adapt. $T$ & 17.94 & 40.15 & 33.90 & 64.46 & \cellcolor{gray!25}{36.52}\\
    \midrule
    \midrule
    \multicolumn{5}{l}{\textbf{Context Reasoning}}\\
    \midrule
    LLaVa 1.5M & 16.11 & 42.69 & 65.64 & 75.08 & \cellcolor{gray!25}{41.47} \\
    ~ + Finetuning & & & & & \cellcolor{gray!25}\\
    ~~~~ {\tiny \cite{chou2024mm}} &
    \multirow{-2}{*}{63.93} &
    \multirow{-2}{*}{76.62} &
    \multirow{-2}{*}{75.00}  & 
    \multirow{-2}{*}{83.91} &
    \multirow{-2}{*}{\cellcolor{gray!25}{\textbf{74.58}}} \\
    ~ + Adapt. $T$ & 31.04 & 55.14 & 72.11 & 81.9 & \cellcolor{gray!25}{55.26} \\
    \bottomrule
\end{tabular}
\label{tab:finetune}
\end{center}
\end{table}

\begin{table}[!t]
\scriptsize
\begin{center}
\caption{\textbf{Ablation Studies on contribution of different loss functions we use in our approach}}
\vspace{-0.5em}
\begin{tabular}{c|c|c|ccccc}
    \toprule
   & $\mathcal{L}_{CE}$ & $\mathcal{L}_{PL}$ & $\mathbf{Acc}$ & $\mathbf{S_{GT}}$ & $\mathbf{Con}$ & $\mathbf{S_{C}}$ & $O_{all}$ \\
    \midrule
    \parbox[t]{4mm}{\multirow{4}{*}{\rotatebox[origin=c]{90}{\parbox{1.15cm}{\centering\textbf{Question \\Rephrasing}}}}} & & & 61.44 & 69.71 & 52.29 & 66.86 & \cellcolor{gray!25}{62.43}\\
    & \checkmark & & 59.48 & 71.70 & 52.94 & 66.36 & \cellcolor{gray!25}{62.48} \\
    & & \checkmark & 66.67 & 76.21 & 85.62 & 88.90 & \cellcolor{gray!25}{78.56}\\
    & \checkmark & \checkmark & 66.01 & 77.18 & 90.20 & 93.10 & \cellcolor{gray!25}{\textbf{80.39}}\\
    \midrule
    \midrule
    \parbox[t]{4mm}{\multirow{4}{*}{\rotatebox[origin=c]{90}{\parbox{1.15cm}{\centering\textbf{Image \\Restyling}}}}} & & & 14.16 & 38.36 & 54.33 & 70.77 & \cellcolor{gray!25}{36.99} \\
    & \checkmark & & 16.12 & 39.86 & 61.86 & 74.10 & \cellcolor{gray!25}{39.65}\\
    & & \checkmark & 17.93 & 40.73 & 83.97 & 89.86 & \cellcolor{gray!25}{43.86}\\
    & \checkmark & \checkmark & 19.25 & 40.35 & 84.94 & 90.40 & \cellcolor{gray!25}{\textbf{44.48}}\\
    \midrule
    \midrule
    \parbox[t]{4mm}{\multirow{4}{*}{\rotatebox[origin=c]{90}{\parbox{1.15cm}{\centering\textbf{Context \\Reasoning}}}}} & & & 32.68 & 27.23 & 31.37 & 55.81 & \cellcolor{gray!25}{35.51} \\
    & \checkmark & & 28.10 & 52.19 & 55.56 & 71.80 & \cellcolor{gray!25}{49.25} \\
    & & \checkmark & 32.77 & 56.13 & 97.14 & 97.1 & \cellcolor{gray!25}{60.99}\\
    & \checkmark & \checkmark & 33.33 & 55.76 & 98.69 & 99.26 & \cellcolor{gray!25}{\textbf{61.44}}\\
    \bottomrule
\end{tabular}
\label{tab:ablation}
\end{center}
\end{table}

\begin{table}[!t]
\scriptsize
\begin{center}
\caption{\textbf{Hyper-parameter search on LLaVa-Next.}}
\vspace{-0.5em}
\begin{tabular}{c|c|c|cccccc}
    \toprule
    & $\alpha$ & $\beta$ & $\mathbf{Acc}$ & $\mathbf{S_{GT}}$ & $\mathbf{Con}$ & $\mathbf{S_{C}}$ & $O_{all}$ \\
    \midrule
    \parbox[t]{4mm}{\multirow{5}{*}{\rotatebox[origin=c]{90}{\parbox{1.3cm}{\centering\textbf{Question \\Rephrasing}}}}} & 0.1 & 1 & 66.01 & 76.38 & 86.27 & 89.81 & \cellcolor{gray!25}{78.73}\\
    &0.5 & 1 & 66.01 & 77.18 & 90.20 & 93.10 & \cellcolor{gray!25}{\textbf{80.39}}\\
    & 1 & 1 & 64.71 & 75.63 & 83.00 & 87.82 & \cellcolor{gray!25}{77.04}\\
    & 1 & 0.5 & 64.71 & 75.68 & 83.01 & 87.88 & \cellcolor{gray!25}{77.07}\\
    & 1 & 0.1 & 65.36 & 75.61 & 79.08 & 84.83 & \cellcolor{gray!25}{75.79}\\
    \midrule
    \midrule
    \parbox[t]{4mm}{\multirow{5}{*}{\rotatebox[origin=c]{90}{\parbox{1.3cm}{\centering\textbf{Image \\Restyling}}}}}& 0.1 & 1 & 17.91 & 40.23 & 86.22 & 91.01 & \cellcolor{gray!25}{43.78} \\
    & 0.5 & 1 & 19.25 & 40.35 & 84.94 & 90.4 & \cellcolor{gray!25}{\textbf{44.48}}\\
    & 1 & 1 & 17.93 & 40.28 & 85.24 & 90.13 & \cellcolor{gray!25}{43.70} \\
    & 1 & 0.5 & 17.93 & 40.28 & 85.26 & 90.47 & \cellcolor{gray!25}{43.73} \\
    & 1 & 0.1 & 17.84 & 40.60 & 82.37 & 87.19 & \cellcolor{gray!25}{43.46}\\
    \midrule
    \midrule
    \parbox[t]{4mm}{\multirow{5}{*}{\rotatebox[origin=c]{90}{\parbox{1.3cm}{\centering\textbf{Context \\Reasoning}}}}} & 0.1 & 1 & 32.68 & 55.32 & 97.39 & 98.11 & \cellcolor{gray!25}{60.68}\\
    & 0.5 & 1 & 33.33 & 55.76 & 98.69 & 99.26 & \cellcolor{gray!25}{\textbf{61.44}}\\
    & 1 & 1 & 33.33 & 55.69 & 97.39 & 98.30 & \cellcolor{gray!25}{61.19}\\
    & 1 & 0.1 & 32.68 & 55.17 & 96.08 & 97.25 & \cellcolor{gray!25}{60.4}\\
    & 1 & 0.5 & 32.68 & 55.21 & 94.77 & 96.30 & \cellcolor{gray!25}{60.20}\\
    \bottomrule
\end{tabular}
\label{tab:hyper}
\end{center}
\end{table}

\begin{figure}[!t]
  \centering
  \includegraphics[width=.5\textwidth]{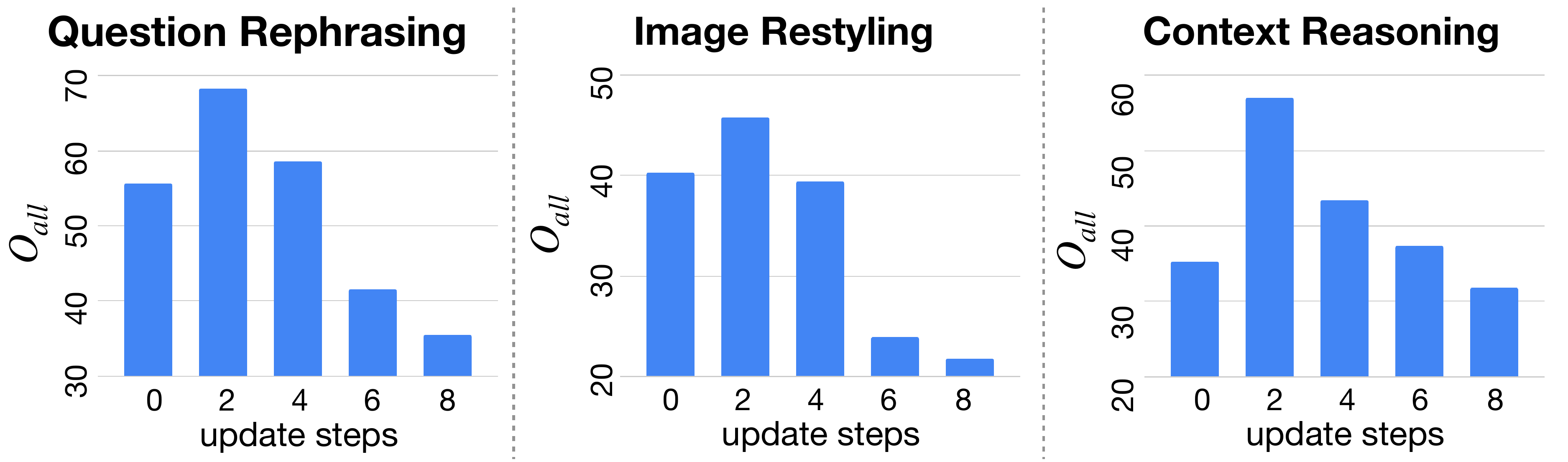}
  \caption{
  We show effect of different number of update steps for each task. 
  }
  \label{fig:updates}
\end{figure}

\subsection{Comparison with Fine-Tuning}

To contextualize the effectiveness of our approach, we compare it against the fully fine-tuned model from MM-R$^3$~\cite{chou2024mm}, which retrains LLaVA-1.5M through large-scale supervised training using task-specific data from the curated MM-R$^3$ training set. In contrast, our method adapts the model using only a single test point and two test-time gradient steps, without access to labeled data, training code, or model internals.

Table~\ref{tab:finetune} presents the results on three MM-R$^3$ tasks. Despite being significantly lighter in terms of computational cost and supervision, our method achieves competitive—and in some cases superior—performance compared to full fine-tuning. Specifically, on the \textit{Question Rephrasing} task, it achieves an $O_{all}$ score of 64.08, outperforming the fine-tuned model (62.02) by a notable margin. 

On \textit{Context Reasoning}, although full fine-tuning achieves the highest score (74.58), our method still improves substantially over the base model (55.26 vs.\ 41.47), again without any retraining. For \textit{Image Restyling}, our method narrows the gap considerably (36.52 vs.\ 46.11), demonstrating strong robustness to visual perturbations even without additional training data. Notably, our method underperforms on these tasks in overall score because full fine-tuning jointly learns the novel task (unsupported by the base VLM) through curated training dataset and improves consistency. It can be seen that the performance of our approach on consistency (i.e Con) is nearly equivalent to that of full-finetuning, while on accuracy the improvement drops. This is not surprising as~\cite{chou2024mm} learns from voluminous training data, which our model is not designed to do being a test-time approach. 

\subsection{Ablation Study}
All experiments in the ablation studies are performed on the LLaVA-Next model, unless specified otherwise.
\subsubsection{Contribution of each component in our test-time consistency framework}
To understand the contribution of each component in our test-time consistency framework, we perform an ablation study by selectively enabling the Cross-Entropy Agreement Loss ($\mathcal{L}_{\mathrm{CE}}$) and the Pseudo-Label Consistency Loss ($\mathcal{L}_{\mathrm{PL}}$). Table~\ref{tab:ablation} reports results across all three MM-R$^3$ tasks.

\paragraph{Complementary Benefits.}  
We observe that both losses independently contribute to improving consistency and overall performance. Applying only $\mathcal{L}_{\mathrm{CE}}$ improves consistency over the base model in all tasks, while $\mathcal{L}_{\mathrm{PL}}$ alone often yields stronger gains in ${\text{Acc}}$. 

\paragraph{Best Results with Combined Loss.}  
The full method—using both $\mathcal{L}_{\mathrm{CE}}$ and $\mathcal{L}_{\mathrm{PL}}$—achieves the highest overall performance across all tasks. For instance, in the question rephrasing task, the combination yields $O_{all} = 80.39$ and consistency of 90.20, outperforming both individual losses. Similar trends are observed in image restyling and context reasoning, where the joint objective achieves the best $O_{all}$ scores of 44.48 and 61.44, respectively. These results show the complementary roles of two losses: $\mathcal{L}_{\mathrm{CE}}$ promotes token-level alignment of outputs across input perturbations, while $\mathcal{L}_{\mathrm{PL}}$ anchors model predictions to a consensus output.

\subsubsection{Ablation on number of updated steps.}

Figure~\ref{fig:updates} shows the impact of varying the number of gradient update steps ($T$) in our \textbf{Test-time (constant $T$)} variant, where a fixed number of updates is applied to all test inputs. We observe that performance improves initially but degrades beyond a certain point, revealing a trade-off between effective adaptation and overfitting. Across all three tasks, setting $T=2$ yields the best score ($O_{all}$).

The performance drop beyond $T=2$ is most pronounced in the \textit{Question Rephrasing} and \textit{Context Reasoning} tasks, likely due to over-adaptation and overfitting on linguistic variations or ambiguous inputs. In contrast, the \textit{Image Restyling} task is relatively robust to the number of updates, suggesting greater stability under visual perturbations.

This ablation is specific to the \textbf{Test-time (constant $T$)} setup. Our alternative variant, \textbf{Test-time (adapt. $T$)}, automatically selects the optimal number of updates per instance using the adaptive step selection mechanism described in Section \ref{Inference}. As such, it does not require manual tuning or per-task sensitivity analysis. Together, these two variants allow us to assess the trade-offs between simplicity and input-specific adaptability.

\vspace{-.5em}
\subsubsection{Ablation on Loss Weighting Coefficients}

We ablate the loss weighting coefficients $\alpha$ and $\beta$ in our total loss $\mathcal{L}_{\text{total}} = \alpha \cdot \mathcal{L}_{\mathrm{CE}} + \beta \cdot \mathcal{L}_{\mathrm{PL}}$, using LLaVA-Next across the MM-R$^3$ tasks. Results in Table~\ref{tab:hyper} show that our method is robust to a range of settings, but performance is highest when both objectives are appropriately balanced.

The best results are obtained with $\alpha = 0.5$ and $\beta = 1$, yielding top $O_{all}$ scores across all tasks: 80.39 (Rephrasing), 44.48 (Restyling), and 61.44 (Reasoning). Performance drops slightly when either loss dominates—for example, using $\beta = 0.1$ reduces consistency and overall score.

\subsubsection{Preservation of Base Model Capabilities}

To ensure that our test-time consistency framework does not degrade the model's original capabilities, we evaluate performance on the unperturbed OKVQA dataset~\cite{okvqa} before and after adaptation. For this experiment, we generate three semantically equivalent rephrasings of each original question using GPT-4V. These rephrasings are used during adaptation, while the final evaluation is performed on the original (unmodified) question from OKVQA dataset.

Results are shown in Table~\ref{tab:original_task}. Both LLaVA-Next and Qwen2-VL improve in accuracy on original unperturbed input after test-time adaptation—rising from 54.69 to 56.10 and from 54.13 to 58.61, respectively. This indicates that our method not only preserves but can even enhance model performance on standard benchmarks. LLaVA 1.5M shows a minor drop (55.09 $\rightarrow$ 53.98), suggesting slightly higher sensitivity in smaller models.
Overall, these results show that our approach does not degrade on the original task distribution, and instead enables consistency improvements.

\begin{table}[!t]
\scriptsize
\begin{center}
\caption{\textbf{Results on original OKVQA dataset task.}}
\vspace{-0.5em}
\begin{tabular}{l|ccc}
    \toprule
    Acc & LLaVA 1.5M & LLaVA-Next & Qwen2-VL \\
     \midrule
     Original & 55.09 & 54.69 & 54.13\\
     ~ + Constant $T$ & 53.98 & 56.10 & 58.61\\
    \bottomrule
\end{tabular}
\vspace{-2.5em}
\label{tab:original_task}
\end{center}
\end{table}
\subsubsection{Ablation on Decoding Temperature.} The ablation studies on different decoding temperatures, $\tau$ = 0, 0.5, 1 are shown in Appendix~\ref{appdex:temp}.
\subsubsection{Qualitative Results.} We show qualitative results for three tasks in the Appendix (see Appendix~\ref{appdex:qualitative} for more details).

\section{Conclusion.}
We present a simple yet effective \emph{test-time consistency} framework for vision–language models that requires no access to curated training data, model internals, or supervised fine-tuning. By leveraging semantically equivalent variants of each input and enforcing agreement through two lightweight losses, our method seamlessly adapts VLMs at inference-time using inherent information in single test-input. Experiments on the MM-R$^3$ benchmark show that our approach significantly improves consistency while preserving or enhancing accuracy. We advocate for consistency as a core evaluation criterion for building reliable, real-world VLM systems in future work.

\vspace{0.05in}
\noindent
{\bf Acknowledgments.}
This work was funded, in part, by the Vector Institute for AI, Canada CIFAR AI Chairs, NSERC Canada Research Chair (CRC), and NSERC Discovery Grants. Resources used in preparing this research were provided, in part, by the Province of Ontario, the Government of Canada through CIFAR, the Digital Research Alliance of Canada\footnote{\url{ https://vectorinstitute.ai/\#partners}}, companies sponsoring the Vector Institute, and Advanced Research Computing at the University of British Columbia. Additional hardware support was provided by John R. Evans Leaders Fund CFI grant and Compute Canada under the Resource Allocation Competition award.

\section*{Limitations}
Our analysis is limited by the scope of the MM-R$^3$ dataset and its predefined perturbations, which may not fully capture the diversity of real-world consistency challenges. While our method improves consistency without access to training data or model internals, it requires multiple forward and backward passes per test input, which increases inference-time latency. However, it remains significantly more efficient and scalable overall compared to full fine-tuning, as it avoids large-scale training and need for supervision. Additionally, since adaptation is performed locally on a single test point, it may not correct broader model deficiencies or systematic biases. Finally, because our approach updates model parameters during inference, it may not be suitable for deployment in strictly frozen or closed-source model environments.

\bibliography{custom}

\clearpage
\appendix
\section{Appendix}
\label{sec:appendix}
\subsection{Ablation on Decoding Temperature}\label{appdex:temp}
\begin{table}[!h]
\scriptsize
\begin{center}
\caption{\textbf{Different temperature on LLaVa-Next.}}
\vspace{-0.5em}
\begin{tabular}{l|cccccc}
    \toprule
    $\mathbf{Temp}$ & $\mathbf{Acc}$ & $\mathbf{S_{GT}}$ & $\mathbf{Con}$ & $\mathbf{S_{C}}$ & $O_{all}$ \\
    \midrule
    \multicolumn{5}{l}{\textbf{Question Rephrasing}}\\
    \midrule
    LLaVa-NEXT, $\tau = 0$ & 42.89 & 64.89 & 49.18 & 65.69 &  \cellcolor{gray!25}{55.61} \\
    ~ + Constant $T$ & 44.48 & 68.74 & 83.39 & 88.47 & \cellcolor{gray!25}{\textbf{68.25}}\\
    LLaVa-NEXT, $\tau = 0.5$ & 42.06 & 65.29 & 52.02 & 66.52 & \cellcolor{gray!25}{56.33}\\
    ~ + Constant $T$ & 44.48 & 68.74 & 83.39 & 88.47 & \cellcolor{gray!25}{\textbf{68.25}}\\
    LLaVa-NEXT, $\tau = 1$ & 42.06 & 65.29 & 52.02 & 66.52 & \cellcolor{gray!25}{56.33}\\
    ~ + Constant $T$ & 44.48 & 68.74 & 83.39 & 88.47 & \cellcolor{gray!25}{\textbf{68.25}}\\
    \midrule
    \midrule
    \multicolumn{5}{l}{\textbf{Image Restyling}}\\
    \midrule
    LLaVa-NEXT, $\tau = 0$ & 17.57 & 41.47 & 55.34 & 71.36 & \cellcolor{gray!25}{40.27}\\
    ~ + Constant $T$ & 18.99 & 42.49 & 88.25 & 91.25 & \cellcolor{gray!25}{\textbf{45.80}} \\
    LLaVa-NEXT, $\tau = 0.5$ & 17.57 & 41.47 & 55.34 & 71.36 & \cellcolor{gray!25}{40.27}\\
    ~ + Constant $T$ & 17.64 & 40.64 & 82.80 & 76.64 & \cellcolor{gray!25}{\textbf{42.68}}\\
    LLaVa-NEXT, $\tau = 1$ & 17.57 & 41.47 & 55.34 & 71.36 & \cellcolor{gray!25}{40.27}\\
    ~ + Constant $T$ & 17.64 & 40.64 & 82.80 & 76.64 & \cellcolor{gray!25}{\textbf{42.68}}\\
    \midrule
    \midrule
    \multicolumn{5}{l}{\textbf{Context Reasoning}}\\
    \midrule
    LLaVa-NEXT, $\tau = 0$ & 30.24 & 27.43 & 32.11 & 58.44 & \cellcolor{gray!25}{35.23} \\
    ~ + Constant $T$ & 32.50 & 50.84 & 89.91 & 90.16 & \cellcolor{gray!25}{\textbf{56.97}} \\
    LLaVa-NEXT, $\tau = 0.5$ & 30.07 & 51.99 & 52.09 & 66.68 & \cellcolor{gray!25}{48.53}\\
    ~ + Constant $T$ & 32.31 & 53.84 & 93.4 & 95.31 & \cellcolor{gray!25}{\textbf{59.15}}\\
    LLaVa-NEXT, $\tau = 1$ & 30.07 & 51.99 & 52.09 & 66.68 & \cellcolor{gray!25}{48.53}\\
    ~ + Constant $T$ & 32.31 & 53.84 & 93.4 & 95.31 & \cellcolor{gray!25}{\textbf{59.15}}\\
    \bottomrule
\end{tabular}
\label{tab:temp}
\end{center}
\end{table}

We conduct an ablation to assess the impact of decoding temperature \( \tau \) on our test-time consistency framework using LLaVA-NEXT across three perturbation types: \textit{Question Rephrasing}, \textit{Image Restyling}, and \textit{Context Reasoning} (Table~\ref{tab:temp}). 

Across all perturbations, our method improves consistency and overall robustness regardless of the temperature setting. Notably:
\vspace{-.3em}
\begin{itemize}[leftmargin=*,noitemsep]
\vspace{-.5em}
    \item \textbf{Question Rephrasing:} Our test-time strategy consistently boosts performance to a peak \( O_{all} = \textbf{68.25} \) for all values of \( \tau \), indicating stable performance across decoding scales and strong resilience to linguistic variations.
    
    \item \textbf{Image Restyling:} While baseline performance is lower due to visual perturbations, our method still yields significant improvements. The best result is observed at \( \tau = 0 \), where \( O_{all} \) improves from 40.27 to \textbf{45.80}, a gain of 5.5 points.
    
    \item \textbf{Context Reasoning:} This task benefits most from our consistency framework. The best performance, \( O_{all} = \textbf{59.15} \), is achieved at both \( \tau = 0.5 \) and \( \tau = 1 \), indicating that our method improves reasoning-heavy tasks.
\end{itemize}

These results demonstrate that our approach is robust to temperature variation and consistently enhances consistency and semantic alignment across all perturbation categories.

\begin{figure*}[!t]
  \centering
  \includegraphics[width=\textwidth]{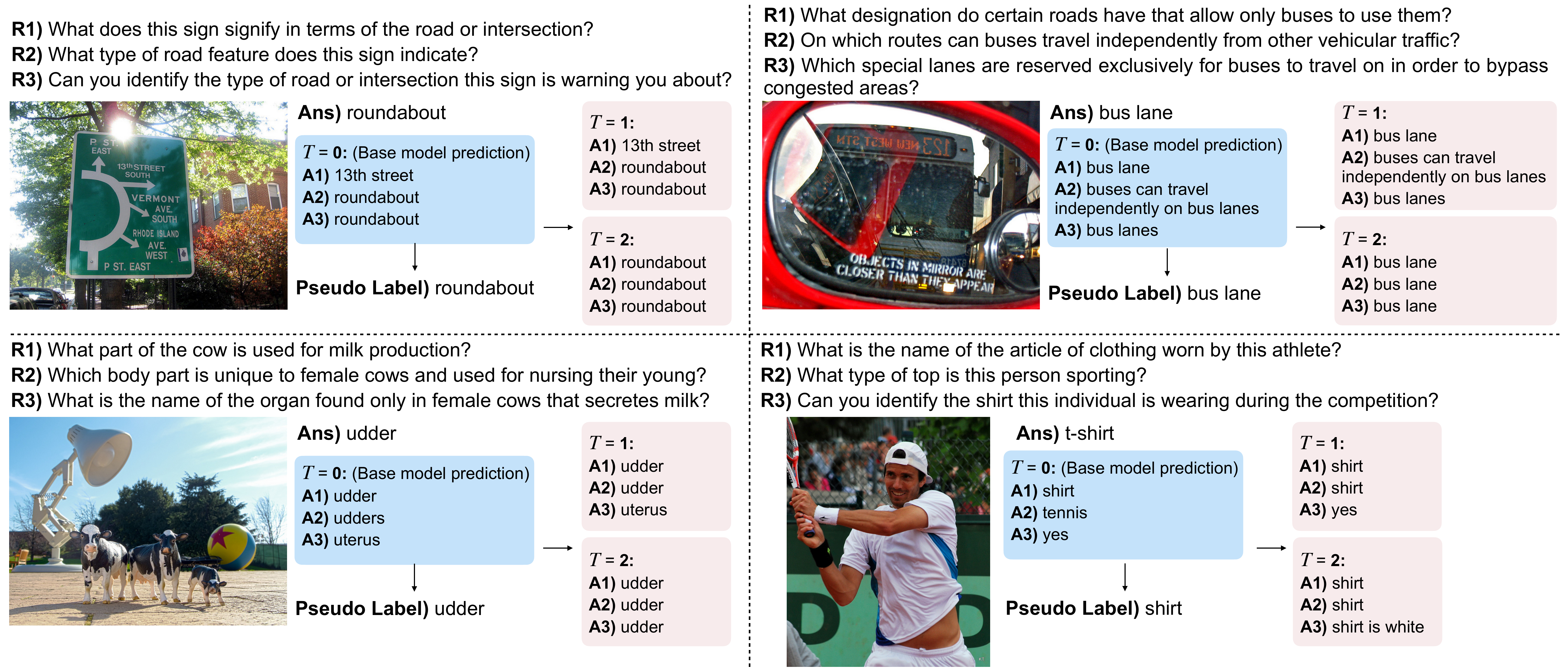}
  \caption{
  Qualitative results on the question rephrasing task.
  }
  \label{fig:qualitative_rephrase}
\end{figure*}

\begin{figure*}[!t]
  \centering
  \includegraphics[width=\textwidth]{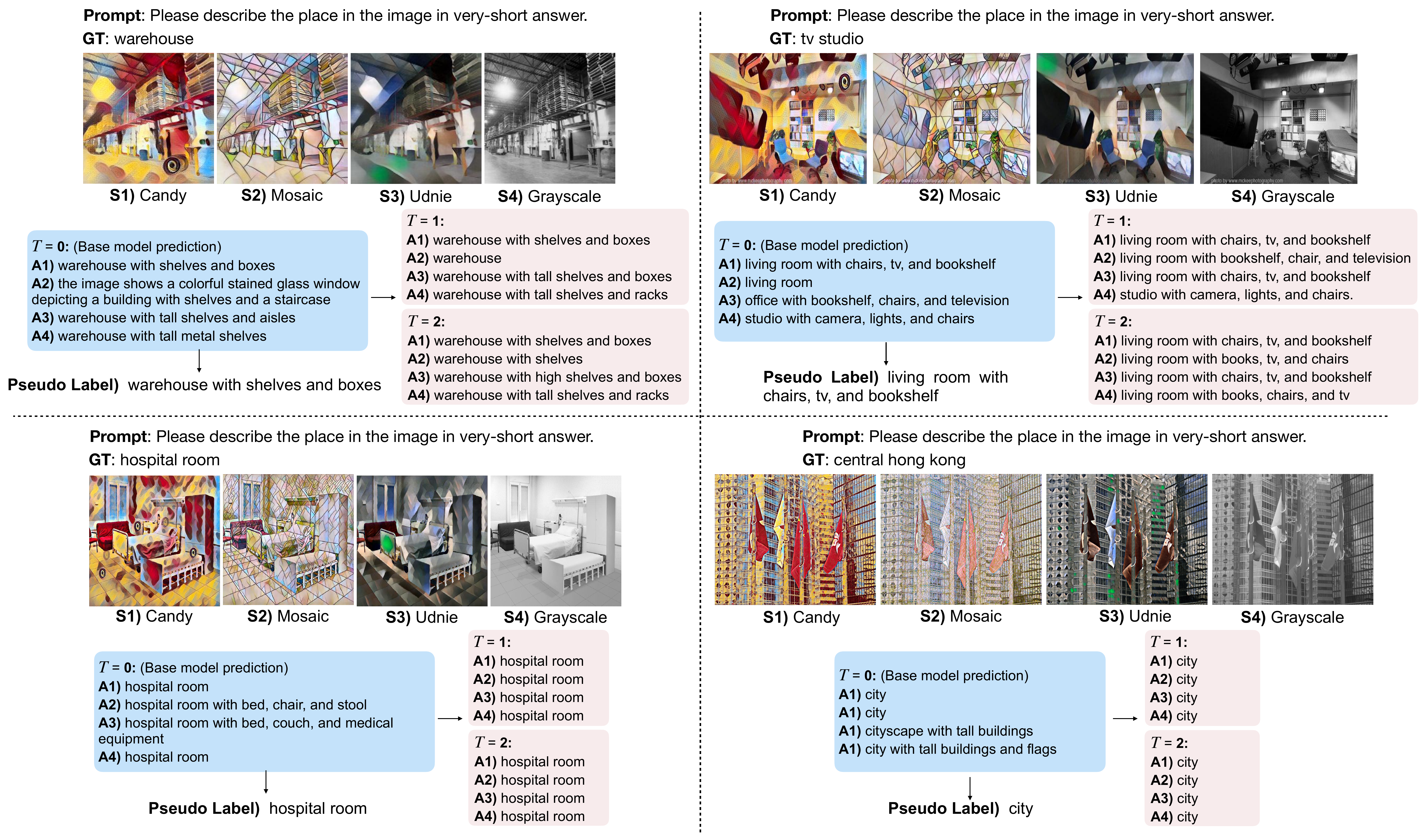}
  \caption{
  Qualitative results on the image restyling task.
  }
  \label{fig:qualitative_style}
\end{figure*}

\begin{figure*}[!t]
  \centering
  \includegraphics[width=.8\textwidth]{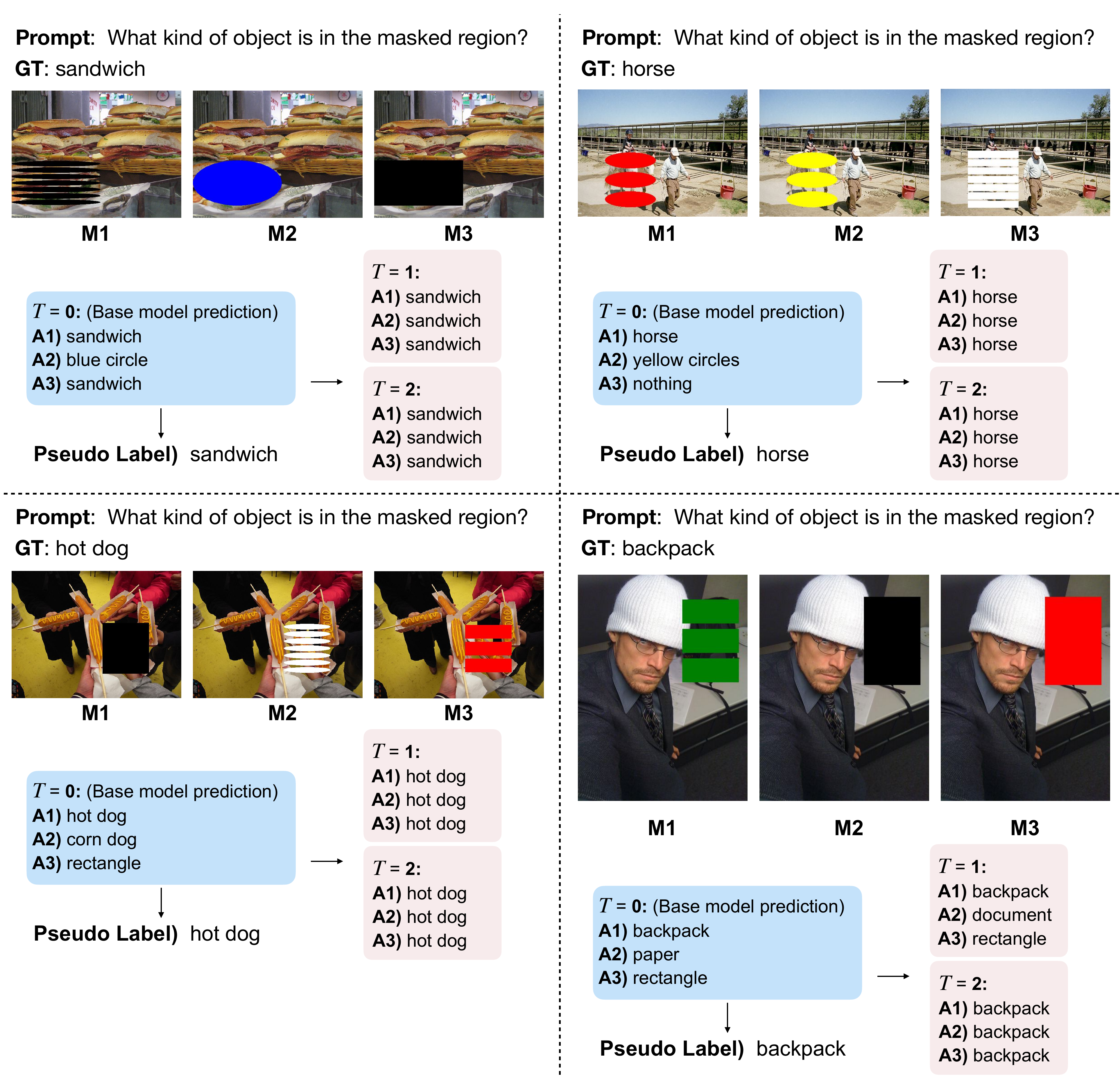}
  \caption{
  Qualitative results on the context reasoning task.
  }
  \label{fig:qualitative_masking}
\end{figure*}

\subsection{Evaluation Metrics}
To systematically assess the performance of VLMs, we use four distinct evaluation metrics, on similar lines as previous work \cite{chou2024mm}, each capturing different aspects of model performance. 

\vspace{0.3em}
\noindent{\textbf{Accuracy ($\mathbf{Acc}$).}}
To evaluate accuracy we 
assess the responses from VLMs based on an fuzzy string matching  with the ground truth annotations, accounting for minor lexical variations. A similarity threshold of 85 is used to determine a match.
The accuracy score is then calculated as the average of correct responses across the benchmark test-set. 
\noindent{\textbf{Similarity with GT ($\mathbf{S_{GT}}$).}}
Given the limitations of exact match criteria—which may penalize semantically correct responses for minor lexical differences—we employ a semantic similarity metric to better evaluate alignment between model outputs and ground truth. For example, terms like \textit{person}, \textit{man}, and \textit{woman} are semantically related but would be treated as mismatches under strict accuracy metrics. To address this, we use BERT-based Sentence Similarity~\citep{reimers-gurevych-2019-sentence}, which leverages contextual language model encodings to assess the semantic alignment between predictions and reference answers. This metric rewards semantic correctness over surface-form similarity. Final scores are computed as the average similarity across the dataset.

\noindent{\textbf{Consistency Accuracy ($\mathbf{Con}$).}}
This metric quantifies the proportion of responses that exhibit a predefined level of semantic consistency. We compute pairwise similarity scores between outputs using the same semantic similarity metric as in $S_{GT}$, and consider a pair consistent if its similarity exceeds a threshold of 0.7—motivated by observations from the Semantic Textual Similarity benchmark~\citep{cer-etal-2017-semeval}. A response is deemed consistent if it meets this threshold with its paired counterpart.

The final score is calculated as the average proportion of consistent pairs across the dataset, providing an aggregate measure of the model's semantic stability across perturbed inputs.

\noindent{\textbf{Consistency Similarity ($\mathbf{S_C}$).}}
Similar to the Consistency Accuracy metric, this measure computes pairwise semantic similarity scores between responses to assess consistency. However, instead of applying a threshold, we take the average of these similarity scores across the dataset. This provides a more \emph{continuous} assessment of the model’s coherence, capturing fine-grained variations in semantic consistency across perturbed inputs.

\vspace{0.3em}
\noindent{\textbf{Overall Performance ($O_{all}$).}}
We report overall model performance using the harmonic mean ($H_{\text{mean}}$) of correctness and consistency scores. Specifically, we first compute the average of $\mathbf{Acc}$ and $\mathbf{S_{GT}}$ to assess correctness, and the average of $\mathbf{Con}$ and $\mathbf{S_C}$ to assess consistency. These two averages are then combined using the harmonic mean:

\begin{equation}
\small
H_{mean}(mean(\mathbf{Acc},\mathbf{S_{GT}}),mean(\mathbf{Con},\mathbf{S_C})).
\vspace{-1em}
\end{equation}
We use the harmonic mean to balance correctness and consistency, as it penalizes models that perform well on only one aspect, thereby encouraging robust performance across both dimensions.

\subsection{Implementation Details.}\label{appdex:implementation}
We use the pre-trained VLMs as the base models and only fine-tune the language modelling head (LM-head) layer. We set updated steps $T = 2$ for the test-time experiments and maximum updated steps to $T = 4$ for the adaptive test-time experiments. The learning rate is set to $5e^{-4}$. All experiments are conducted on NVIDIA A40 with batch size 1 on all three models.

\subsection{Qualitative Results}\label{appdex:qualitative}
We show qualitative results for the question rephrasing task in Figure~\ref{fig:qualitative_rephrase}, image restyling in Figure~\ref{fig:qualitative_style}, and context reasoning in Figure~\ref{fig:qualitative_masking}. Across all three tasks, even when the base model predictions are inconsistent, our method is able to further improve consistency and thus overall score (as also supported by quantitative results in Main manuscript).

\end{document}